\documentclass[10pt,twocolumn,letterpaper]{article}

\newcommand{\Figure}[1]{Figure~\ref{fig:#1}}
\newcommand{\Table}[1]{Table~\ref{tab:#1}}
\newcommand{\Eq}[1]{Eq.~\eqref{eq:#1}}

\newcommand{\vect}[1]{\ensuremath{\mathbf{#1}}\xspace}
\newcommand{\matr}[1]{\ensuremath{\mathbf{#1}}\xspace}

\newcommand{\SO}[1]{\ensuremath{\mathbb{SO}(#1)}\xspace}
\newcommand{\SE}[1]{\ensuremath{\mathbb{SE}(#1)}\xspace}
\newcommand{\R}[1]{\ensuremath{\mathbb{R}^{#1}}\xspace}

\newcommand{\Rot}{\matr{R}}
\newcommand{\Transl}{\vect{t}}
\newcommand{\Transf}{\matr{M}}

\newcommand{\argmin}{\mathop{\mathrm{arg\,min}}}

\usepackage{gensymb}
\usepackage{lipsum}

\usepackage{array}
\usepackage{hhline}
\usepackage{ctable}
\newcolumntype{C}[1]{>{\centering\let\newline\\\arraybackslash\hspace{0pt}}m{#1}}
\newcolumntype{R}[1]{>{\raggedleft\let\newline\\\arraybackslash\hspace{0pt}}m{#1}}

\usepackage{amsmath}
\usepackage{xspace}

\usepackage{csvsimple}

\newcommand{\X}{\vect{X}}
\newcommand{\Y}{\vect{Y}}
\newcommand{\Z}{\vect{Z}}

\usepackage{cvpr}
\usepackage{times}
\usepackage{epsfig}
\usepackage{graphicx}
\usepackage{amsmath}
\usepackage{amssymb}

\usepackage{booktabs}
\usepackage{csvsimple}
\usepackage{siunitx}
\usepackage[caption=false, font=footnotesize]{subfig}
\usepackage[pagebackref=true,breaklinks=true,letterpaper=true,colorlinks,bookmarks=false]{hyperref}

\cvprfinalcopy 

\def\cvprPaperID{4} 

\ifcvprfinal\fi

\begin{document}

\title{Integration of Absolute Orientation Measurements in the KinectFusion Reconstruction pipeline}

\author{Silvio Giancola, Jens Schneider, Peter Wonka and Bernard S. Ghanem\\
King Abdullah University of Science and Technology (KAUST), Saudi Arabia\\
{\tt\small \{silvio.giancola, jens.schneider, peter.wonka, bernard.ghanem\}@kaust.edu.sa}
}

\maketitle
\thispagestyle{empty}

\begin{abstract}

In this paper, we show how absolute orientation measurements provided by low-cost but high-fidelity IMU sensors can be integrated into the KinectFusion pipeline. 
We show that integration improves both runtime, robustness and quality of the 3D reconstruction.
In particular, we use this orientation data to seed and regularize the ICP registration technique.
We also present a technique to filter the pairs of 3D matched points based on the distribution of their distances. This filter is implemented efficiently on the GPU. 
Estimating the distribution of the distances helps control the number of iterations necessary for the convergence of the ICP algorithm. 
Finally, we show experimental results that highlight improvements in robustness, a speed-up of almost 12\%, and a gain in tracking quality of 53\% for the ATE metric on the Freiburg benchmark.

\end{abstract}

\begin{figure}[t]
	\centering
	{\includegraphics[width=0.245\textwidth]	{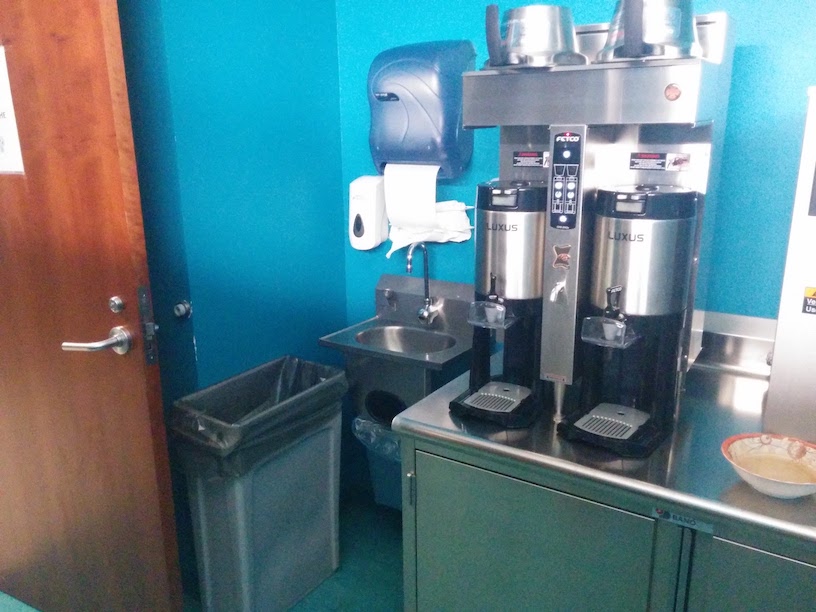}}
	{\includegraphics[width=0.225\textwidth]	{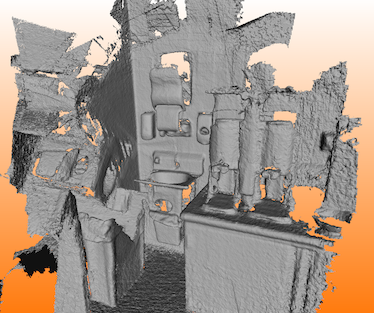}}\\
	\vspace{1mm}
	{\includegraphics[width=0.48\textwidth] {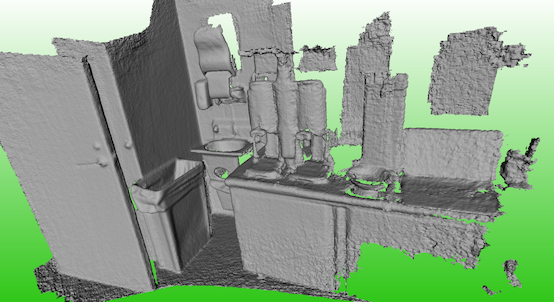}}			
	\caption{Reconstruction of a \textit{Kitchen} environment using an absolute orientation prior to seed the ICP algorithm.
		\textbf{\textit{Top Left}}: Photo of the Kitchen environment.
		\textbf{\textit{Top Right}}: Original KinectFusion reconstruction.
		\textbf{\textit{Bottom}}: IMU-seeded KinectFusion reconstruction.
	}
	\label{fig:Teaser}
\end{figure}%

\section{Introduction}

Automated 3D reconstruction of geometry from images is a highly versatile field of research with applications in 
archeology~\cite{2017arXiv170802033G}, 
topography~\cite{giberti2016accuracy}, 
urban planning~\cite{musialski2013survey}, 
robotics~\cite{konolige2017detection} and 
entertainment~\cite{pagliari2014kinect}.
The ubiquity of this core vision task stems primarily from decades of advances in reconstruction algorithms \cite{nister2005preemptive,thrun2002robotic,durrant2006simultaneous,izadi2011kinectfusion} and acquisition devices \cite{axelsson1999processing,geng2011structured,foix2011lock,khoshelham2012accuracy,smisek20133d,corti2016metrological}. 
With the latest advent of low-cost consumer-class RGB-D cameras, 3D scene understanding and reconstruction has become a pervasive mass market technology.

Simultaneous Localization and Mapping (SLAM) techniques are based on a registration step, i.e., they seek to properly align pairs of  images or point clouds. 
This registration step consists of finding a $6$-DoF rigid body transformation $ \Transf \in \SE3 $ composed of a rotation matrix $\Rot \in \SO3$ and a translation vector $\Transl \in \R3$ such that overlapping parts of a scene can be superimposed, thereby minimizing the distances between pairs of matched points.
It is clear that the process of finding the best-possible rigid transformation $\Transf = (\Rot,\Transl)$ is affected by noise, occlusion, matching quality, and outliers. 
Therefore, it remains intrinsically ambiguous and complex to solve (\Rot,\Transl) in a robust and computationally efficient fashion. 
The main reason is that solving for both \Rot and \Transl simultaneously is a non-convex problem.
We believe that non-convexity is principally introduced by the orientation component.
The problem can be linearized, however, if a rough estimate of the orientation is available.

Point cloud registration, be it frame-by-frame or frame-to-model, is the foundation  for 3D reconstruction, since separate acquisitions have to be aligned.
The original KinectFusion algorithm~\cite{newcombe2011kinectfusion} makes use of a coarse-to-fine ICP method~\cite{besl1992method} with a fixed number of iterations, using a linear approximation~\cite{low2004linear} that minimizes for the point-to-plane metric~\cite{rusinkiewicz2001efficient}.

In this work, we present a modified KinectFusion framework that integrates an absolute orientation measurement provided by an off-the-shelf IMU.
The orientation prior is leveraged in the ICP method and the original problem is solved with a regularization term for the orientation component.  
Also, correspondences between closest points are filtered based on an efficient estimate of the median distance between pairs of points implemented on a GPU. As such, we observe improvements in quality, runtime, and robustness in  the 3D reconstruction process. We characterize the uncertainty model for the orientation measurement according to the ISO JCGM 100:2008~\cite{bipm2008evaluation}, and we apply such model to the Freiburg dataset~\cite{sturm2012benchmark} in order to evaluate the quality of our reconstruction pipeline.


\section{Related work} \label{sec:RelatedWork}

Visual-Inertial SLAM is a hot topic with direct applications to robotics, for which the setup usually includes cameras and IMUs.
Here, we review works focusing on \textit{Navigation} and \textit{Mapping}, dual topics applied to robotics.

\textbf{Navigation} focuses on estimating the system's ego-motion, i.e., its own trajectory in space and time.
Many important navigation methods use \emph{extended Kalman filter} (EKF) techniques~\cite{paul2017comparative, wu2015square, huang2014towards, li2013high}.
They define and update a system state based on visual and inertial measurements using advanced and complex Kalman filtering techniques~\cite{mourikis2007multi,martinelli2012vision,ligorio2013extended}.
Paul et al.~\cite{paul2017comparative} recently conducted a comparative study between monocular and stereo visual-inertial EKF-based techniques, but do not mention any RGB-D devices.
Other methods such as~\cite{leutenegger2015keyframe, forster2015imu, qayyum2013inertial} focus on solving a \emph{global optimization problem} minimizing for a combined cost function.
In those techniques, the translation is estimated by integrating the IMU acceleration measurements twice. 
However, since integration may amplify noise, such double integration generally produces significant drifts.
Leutenegger et al.~\cite{leutenegger2015keyframe} modeled the noise in a probabilistic way, but residual drift inevitably affects the method's stability negatively.
In contrast, we show that orientation measurements can be estimated robustly and consistently using a $9$-DoF IMU.
In Visual-Inertial Navigation techniques, the IMU data can be \emph{loosely-coupled}, i.e., an independent measurement in the problem \cite{konolige2010large,weiss2012real}
or \emph{tightly-coupled}, optimizing over all sensor measurements in the optimization~\cite{leutenegger2015keyframe} and EKF pipelines~\cite{mourikis2007multi,wu2015square}.
Finally, it is worth noting that navigation techniques mostly make use of monocular or stereo cameras, but do not rely on direct depth measurements, mainly because the quality of the 3D reconstruction is usually out-of-scope for navigation tasks.

\textbf{Mapping.}
While Navigation tries to solve for ego-motion, Mapping focuses attention on the 3D reconstruction of the surrounding environment.
Zhang et al.~\cite{zhang2017enabling} recently presented a pipeline that estimates ego-motion while building an accurate representation of the surrounding environment.
Similar to Navigation, it is common practice to rely on EKF~\cite{brunetto2015fusion} or global optimization~\cite{ma2016large} techniques.
Ma et al.~\cite{ma2016large} presented a solution for large-scale Visual-Inertial reconstruction using a volumetric representation similar to KinectFusion.
Concha et al.~\cite{concha2016visual} created a real-time, fully dense reconstruction based on an RGB camera and an IMU.

\textbf{Visual Inertial KinectFusion.}
Nie{\ss}ner et al.~\cite{niessner2014combining} followed similar considerations by using an IMU from a smartphone to initialize both (\Rot,\Transl) in the ICP step of KinectFusion. 
While achieving some improvement in runtime, reliable position information cannot be extracted from an IMU alone, since the accelerometer's noise leads to significant drift. 
Smartphones therefore use a fusion of multiple triangulation techniques based on measurements including GSM, WiFi, and GPS (outdoors). Nevertheless, the position information is of very low fidelity, especially indoors. 
We argue that since ICP is generally sensitive to its initialization, providing it with such noisy measurements can only lead to limited improvement (or even degradation) in many acquisition scenarios. 
In contrast to requiring an expensive, full sensor array as found in smartphones, our method only relies on a cheap and modular IMU that produces high-fidelity absolute orientation measurements in real-time.

\textbf{Datasets.} 
Pfrommer et al.~\cite{pfrommer2017penncosyvio} recently released a dataset for visual-inertial benchmarking. 
However, this dataset  strictly focuses on odometry along long distances, but not on small range reconstructions as KinectFusion-based algorithms do.
Sturm et al.~\cite{sturm2012benchmark} released multiple versions of their well-known Freiburg benchmark, albeit without orientation measurements.
We will present a way to generate such measurements using the ground truth orientation and an empirical uncertainty model of our sensor.

\textbf{Contributions.} 
While the existing literature focuses on solving the problem for both position and orientation from the IMU raw measurements, we argue that orientation alone is sufficient to improve overall performance.
Our contributions are:
\textbf{(i)} We overcome the non-convexity of the joint problem induced by the unknown rotation \Rot by seeding the ICP algorithm with an orientation estimate from the IMU.
\textbf{(ii)} We present a regularized point-to-plane metric in the coarse-to-fine ICP alignment.
\textbf{(iii)} We use a novel filter that estimates the median distance between closest points efficiently on the GPU in order to reject outliers and control ICP convergence.
\textbf{(iv)} We provide an uncertainty model of the  IMU measurements and apply our model to the Freiburg dataset.


\section{Proposed method} 
\label{sec:ProposedMethod}

The original KinectFusion algorithm is based on a coarse-to-fine ICP between a newly acquired frame and the model reconstructed so far. 
The current frame is aligned by solving the non-linear least squares minimization problem shown in \Eq{PointToPlane} using the point-to-plane metric to find the optimal transformation matrix \Transf $\in$ \SE3, composed of an orientation \Rot $\in$ \SO3 and a translation \Transl $\in$ \R3.

\begin{equation}
    \Transf_{opt} = \argmin_{\Transf} \sum_{i}{ \left( \left(\Transf \mathbf{p}_i - \mathbf{q}_i \right) ^\top \mathbf{n}_i\right) ^ 2},
    \label{eq:PointToPlane}
\end{equation}

where $\mathbf{p}_i$ and $\mathbf{q}_i$ are pairs of matched points belonging to the current frame and the model respectively.
In our method, we use the orientation measurement from the IMU to pre-orient the point $\mathbf{p}_i$ beforehand.
Inspired by Low et al.~\cite{low2004linear}, who proposed a linear solution to this problem by assuming small angles ($\alpha\simeq0$, $\beta\simeq0$, $\gamma\simeq0$), transformation \Transf is approximated by $\widehat{\Transf}$ according to \Eq{ApproxTransf}, intended to solve for small angle after applying the IMU seed.

\begin{equation}
    \begin{aligned}
    \Transf \simeq 
    \begin{pmatrix} 
    1 & -\gamma & \beta & t_x \\
    \gamma & 1 & -\alpha & t_y \\
    -\beta & \alpha & 1 & t_z \\
    0 & 0 & 0 & 1 
    \end{pmatrix} := \widehat{\Transf}.
    \end{aligned}
    \label{eq:ApproxTransf}
\end{equation}

In this small orientation setting, the optimization of \Eq{PointToPlane} can be reformulated into the linear least square shown in \Eq{PointToPlaneLLS}, where $\mathbf{x} = (\alpha,\beta,\gamma,t_x,t_y,t_z)^\top$ and $\mathbf{A}$ is a $n\times6$ matrix.
Please refer to the original implementation by Low et al.~\cite{low2004linear} for further details on $\mathbf{A}$ and $\mathbf{b}$.

\begin{equation}
    \mathbf{x}_{opt} = \min_\mathbf{x} \frac{1}{2}\|\mathbf{Ax}-\mathbf{b}\|_2^2
    \label{eq:PointToPlaneLLS}
\end{equation}

To solve \Eq{PointToPlaneLLS}, $\mathbf{A}^\top\mathbf{A}$ and $\mathbf{A}^\top\mathbf{b}$ are efficiently computed on the GPU using a parallel reduction \cite{harris2007parallel}. The resulting 6$\times$6 linear system $\mathbf{A}^\top\mathbf{A}\mathbf{x} = \mathbf{A}^\top\mathbf{b}$ is efficiently solved using a Singular Value Decomposition (SVD) performed on the CPU.

\subsection{Regularized formulation of the problem}

After leveraging the orientation prior, we propose to solve the regularized formulation shown in \Eq{LLS}.

\begin{equation}
    \mathbf{x}_{opt} = \min_{\mathbf{x}} { \frac{1}{2n} \|\mathbf{A}\mathbf{x}-\mathbf{b} \|_2 ^2 } + \lambda \|\mathbf{P}\mathbf{x}\|_2^2.
    \label{eq:LLS}
\end{equation}

We set $ \mathbf{P} = \begin{bmatrix} \mathbf{I}_3 | \mathbf{0}_3   \end{bmatrix}$  to regularize only the 3 angles $\alpha$, $\beta$ and $\gamma$ to be close to zero. Such a constraint enforces the small angle hypothesis presented in the original formulation \eqref{eq:ApproxTransf}, but also allows for small rotations to compensate for the noise in the IMU orientation measurement. The regularized linear system that we solve is presented in Eq.~\eqref{eq:RegulLinearSystem}.

\begin{equation}
(\mathbf{A}^\top\mathbf{A} + 2 \lambda n \mathbf{P}^\top\mathbf{P})\mathbf{x} = \mathbf{A}^\top\mathbf{b}
\label{eq:RegulLinearSystem}
\end{equation} 

The regularized term $2 \lambda n \mathbf{P}^\top\mathbf{P}$ is efficiently computed on the GPU. 
$\lambda$ is a hyper-parameter that leverages the use of the IMU prior; it can be fixed (constant) or a function of $n$. The regularized formulation of the linearized problem is solved using an SVD on the CPU.
The process is repeated in an ICP framework.

\subsection{Distribution of closest-point distances}

In the original problem, the number of \emph{good} matches $n$ in Eq. \eqref{eq:LLS} is unknown.
To both estimate $n$ and identify inconsistent correspondences, we exploit the probability density function (PDF) of closest point pairs, which gathers the point-to-point distances into a histogram. The PDF and its cumulative (CDF) are efficiently computed on the GPU and are used in the next steps. Figure~\ref{fig:MedianFiltering} shows a PDF extracted from a single iteration in the ICP framework.

\subsection{Median-based filtering}
We also propose a filtering method for the closest-point correspondences based on the median distance.
This median  is estimated from the CDF of the distances obtained up to the resolution of the used histogram.
We argue that, by correcting the orientation, the distribution of the closest point distances should be clustered around the distance corresponding to the translation between the two point clouds. In other words, the distances should be gathered around $\bar{d} = \sqrt{t_x^2 + t_y^2 + t_z^2}$ after orientation correction.

\begin{figure}[htb]	
	\centering
	\includegraphics[width=0.5\textwidth]
	{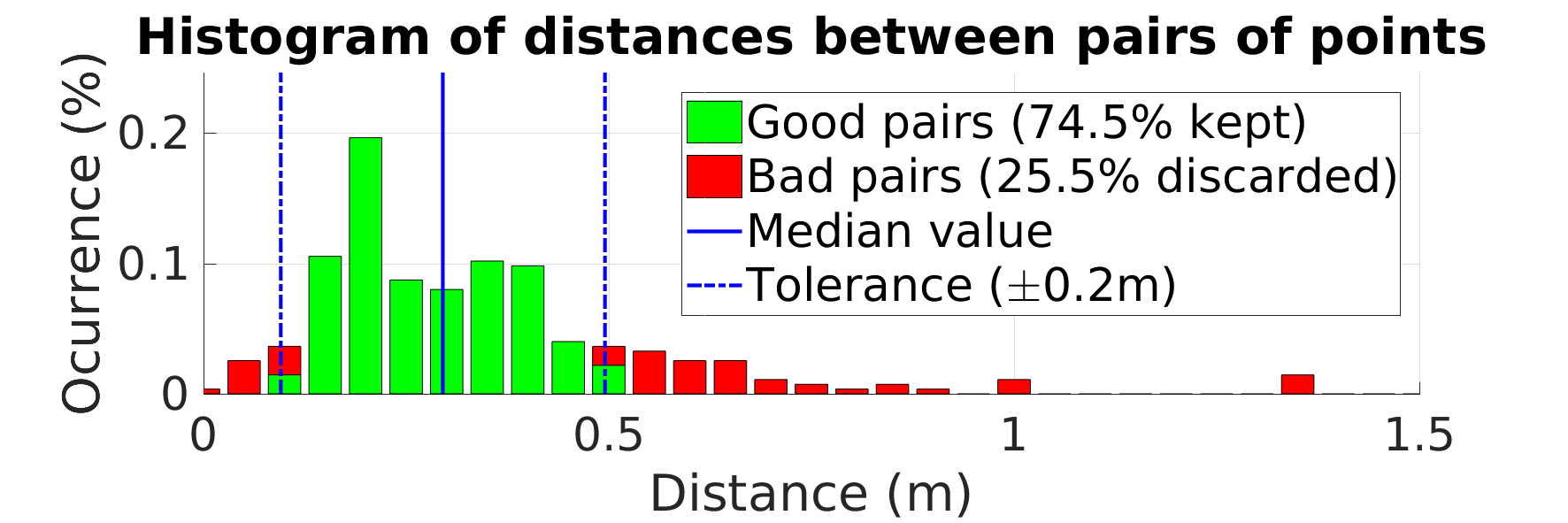}
	\caption{Median Filtering performed on a distance distribution. In this specific case, we are keeping around $75\%$ of the pairs. }
    \label{fig:MedianFiltering}
\end{figure}

The closest point estimation is performed using a normal shooting technique, with normals estimated using an integral image method~\cite{holzer2012adaptive}.
Such a normal estimation is prone to error especially on the border of the depth frame and on the edges of objects in the scene. The algorithm may look for the closest point along a wrong direction.
We thus filter out pairs of points that do not gather around the median distance, identifying them as erroneous.
Figure~\ref{fig:MedianFiltering} illustrates such median-based filtering. Here, the PDF of the distances is plotted with the good pair distances in green and the filtered out ones in red.

\subsection{Convergence verification}

The median distance used for the filtering is subsequently used to control the convergence of the ICP.
Since we seed the ICP with the orientation, we observe that the coarse-to-fine ICP requires less  iterations for convergence than before. To assess the proper alignment, we loop until the median converges instead of performing a fixed number of iterations. The convergence of a given level of the coarse-to-fine ICP is achieved when the median no longer changes within its resolution for 3 successive iterations.


\section{Experiments} \label{sec:Evaluation}

\begin{figure}[htb]	
	\centering
	\includegraphics[height=3.0cm]
	{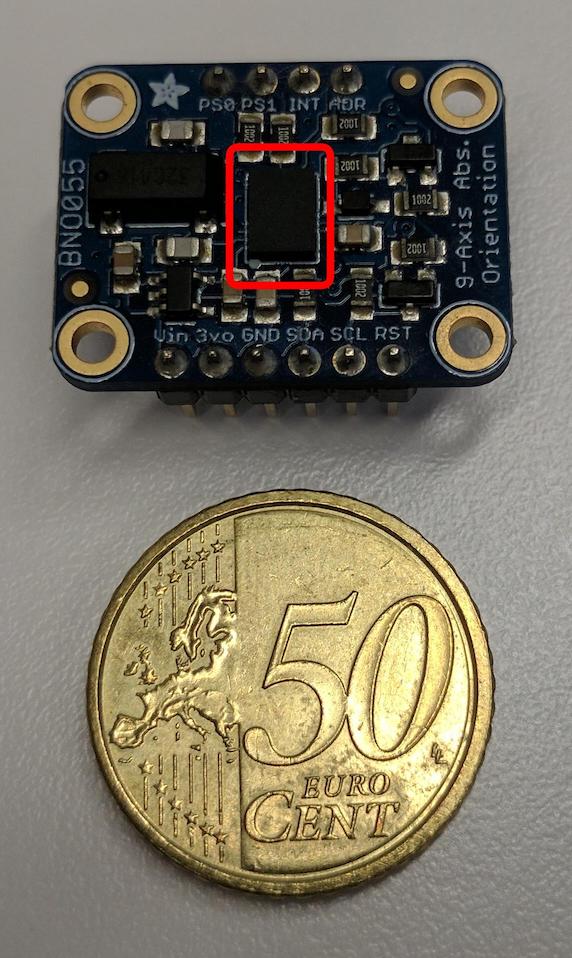}
	\includegraphics[height=3.0cm]    
	{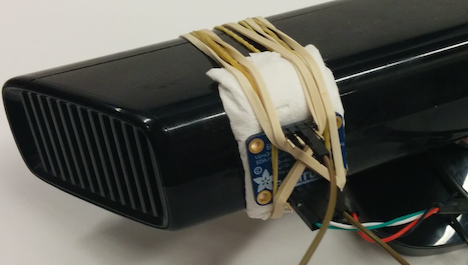}	
	\caption{BNO055 board on a Kinect V1 RGB-D Camera.}
    \label{fig:IMU_Setup}
\end{figure}

We use the state-of-the-art IMU BNO055 manufactured by Bosch, which is typically designed for embedded applications, such as flight control and motion capture. It has a small footprint of a few mm$^2$ and comes on a 1cm-sized board. We mounted the IMU on a Kinect~V1 (Figure~\ref{fig:IMU_Setup} right) and its pose has been calibrated with the reference system of the 3D camera using the hand-to-eye calibration method proposed by Tsai et al.~\cite{tsai1989new}. The Kinect~V1 grabber has been modified to include the IMU orientation in the point cloud. We used the KinectFusion implementation available in the Point Cloud Library \cite{Rusu_ICRA2011_PCL}.

\subsection{Evaluation of ICP Variants}

\begin{figure}[t]
	\centering
	\begin{minipage}[c]{.15\textwidth}
		\subfloat[][Input point cloud.]
		{\includegraphics[width=\textwidth]{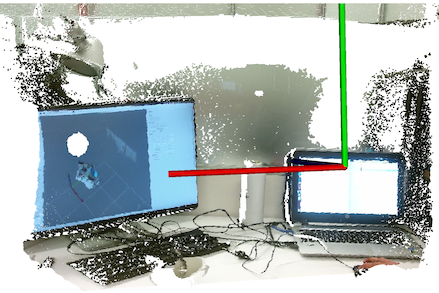}}\\
		\subfloat[][Target point cloud.]
		{\includegraphics[width=\textwidth]{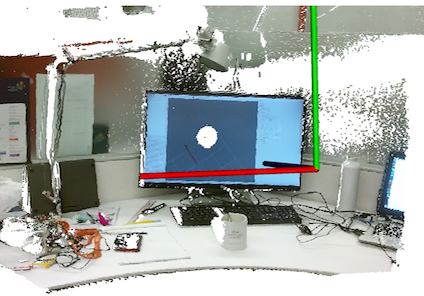}}
	\end{minipage}~ 
	\begin{minipage}[c]{.3\textwidth}    
		\centering
		\subfloat[][ICP convergence over iterations]
		{\includegraphics[width=\textwidth]{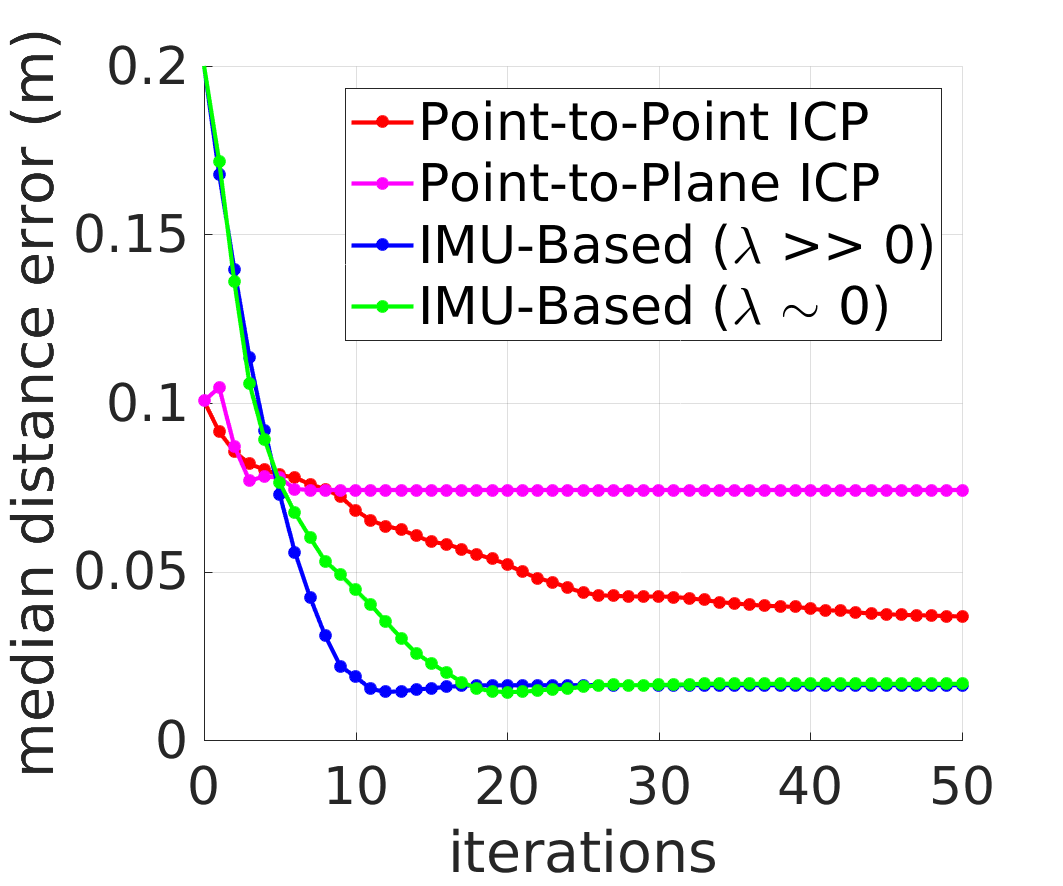}
			\label{fig:convICPplot}}
	\end{minipage}\\
	\caption{Input (a) and target (b) point clouds with IMU measurements, the convergence (c) of ICP metric (median) without IMU vs. IMU-Based methods. }
	\label{fig:ICP3D}
\end{figure}

We tested different versions of ICP using an illustrative \textit{Desktop} example for the registration, shown in Figure~\ref{fig:ICP3D}. Here, we consider two consecutive frames in the \textit{Desktop} sequence and evaluate how well each ICP variant registers the input point cloud to the target point cloud.

From \Figure{convICPplot}, it is evident that ICP may converge to an undesirable local minimum if the IMU is not used, especially if the initial guess of the matched point correspondences is unreliable. 
Here, even though the IMU-Based metrics have a worse initial value, it lies in a more linear and more convex neighbourhood of the global optimal solution, hence it converges faster and more robustly to the desired solution.

\begin{figure}[t]		
	\centering 

	\subfloat[][Original ICP]{
	\includegraphics[width=0.23\textwidth]{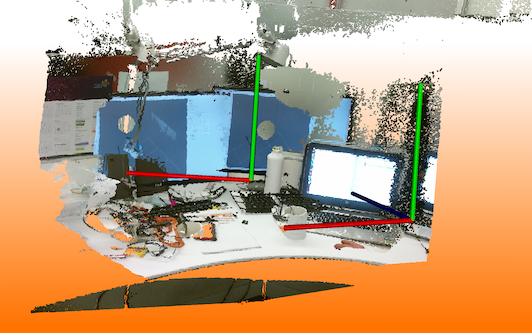}
	\label{fig:ICPNOIMUPC}
	\includegraphics[width=0.23\textwidth,trim={0 0 0 0.5cm},clip]{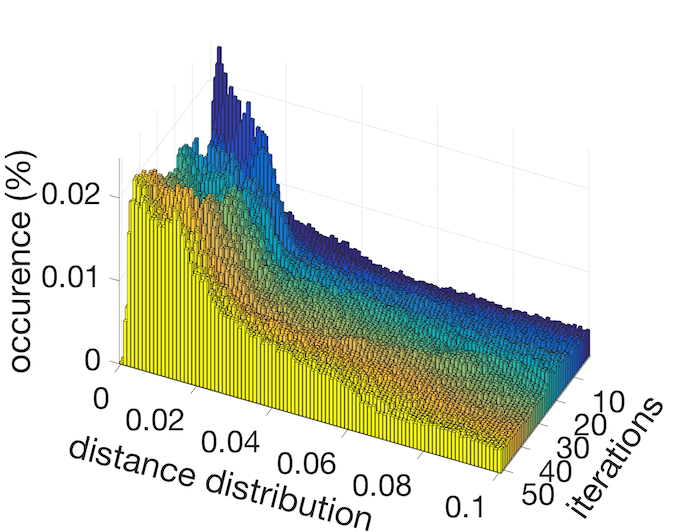}
	\label{fig:ICPNOIMUDist}}\\
	\vskip-0.5mm
	\subfloat[][IMU-Based ICP (strongly regularized $\lambda \gg 0$)]{
	\includegraphics[width=0.23\textwidth]{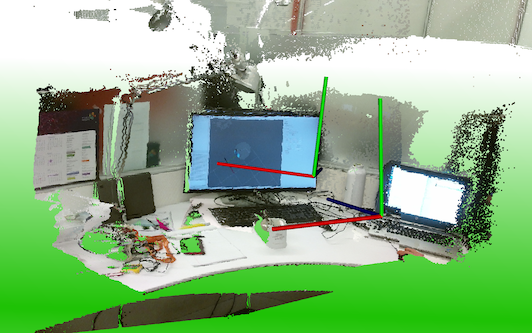}
	\label{fig:ICPIMUBasedPC}				
	\includegraphics[width=0.23\textwidth,trim={0 0 0 0.5cm},clip]{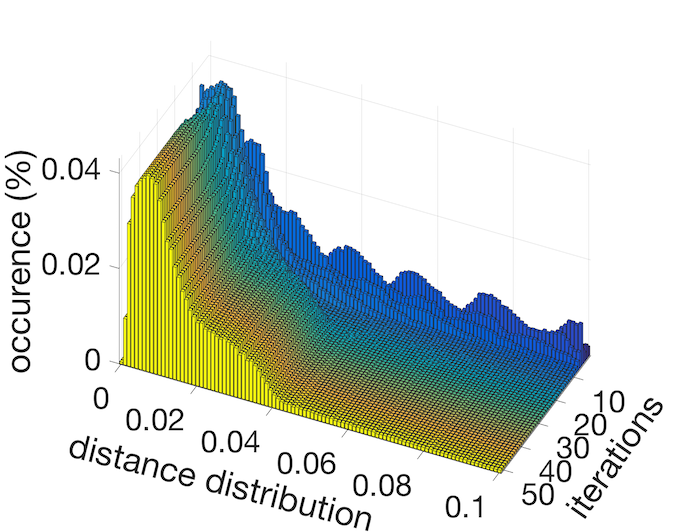}
	\label{fig:ICPIMUBasedDist}}\\   
	\vskip-0.5mm
	\subfloat[][IMU-Based ICP (not regularized $\lambda = 0$)]{
	\includegraphics[width=0.23\textwidth]{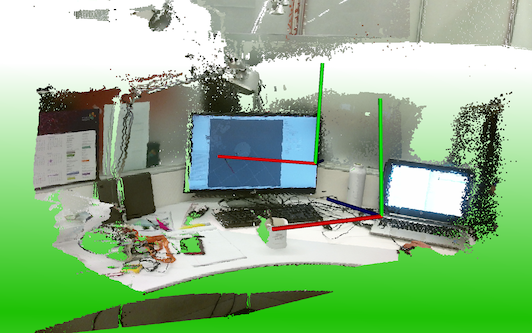}
	\label{fig:ICPIMUInitPC}
	\includegraphics[width=0.23\textwidth,trim={0 0 0 0.5cm},clip]{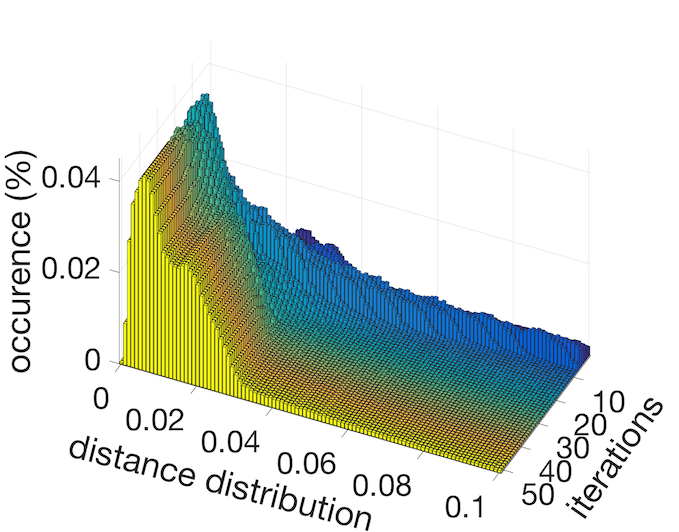}
	\label{fig:ICPIMUInitDist}}\\
	
	\caption{ICP alignment and distance distribution over iteration.}
	\label{fig:ICPNOIMU}
	
\end{figure}

\Figure{ICPNOIMU} compares the results of the alignment of two point clouds, as well as, the distribution of the closest point distance over iterations, using a plain ICP and our IMU-Based ICP both with and without regularization. In this example, the original ICP gets stuck in a local minimum and does not align  the two point clouds properly (\Figure{ICPNOIMUDist}); the distribution of distances does not converge after 50 iterations.
In Figures~\ref{fig:ICPIMUBasedDist} and \ref{fig:ICPIMUInitDist} we show how our system converges when initialized with the orientation measurement. 
Using a strong regularization (b), the orientation is not optimized, the distribution does converge, but, in this case, the translation \Transl is computed without taking into account the eventual noise in the orientation.  
Without regularization (c), the robustness is decreased, i.e. the ICP can converge to a local minimum, but an eventual error in the orientation measurement is corrected, which produces a narrower PDF. 



\subsection{Angular convergence}

\begin{figure}[t]
	
	\centering
	
	\includegraphics[width=0.23\textwidth]
	{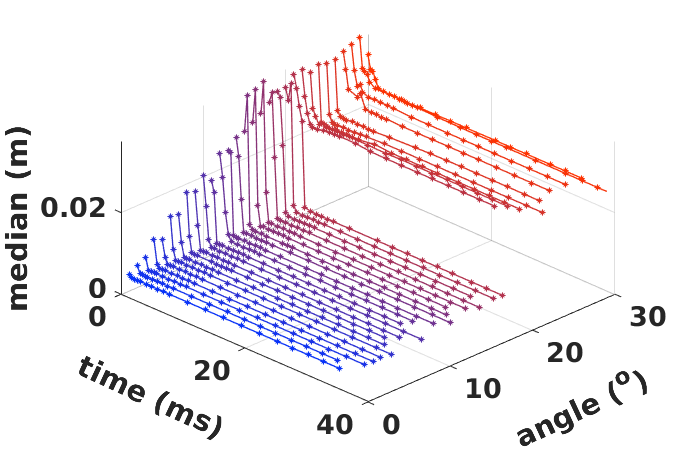}
	\includegraphics[width=0.24\textwidth]
	{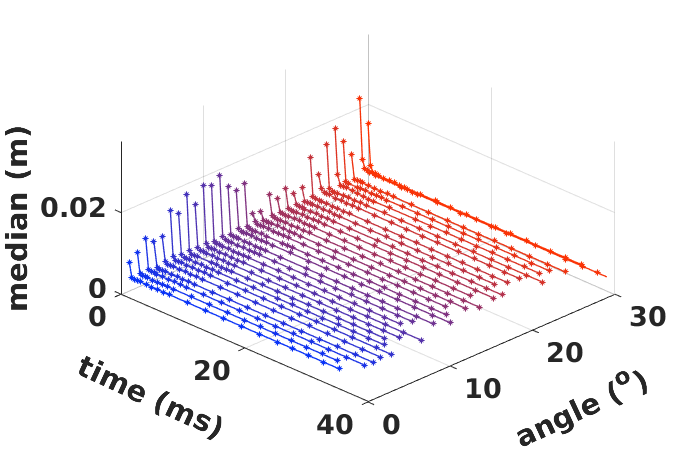}\\

	\includegraphics[width=0.23\textwidth]
	{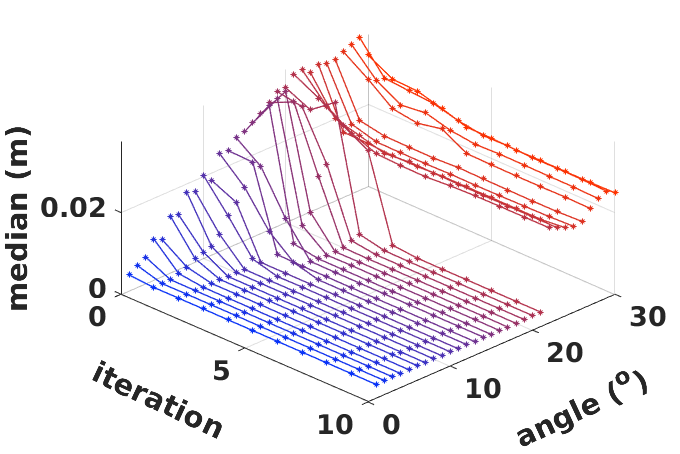}
	\includegraphics[width=0.24\textwidth]
	{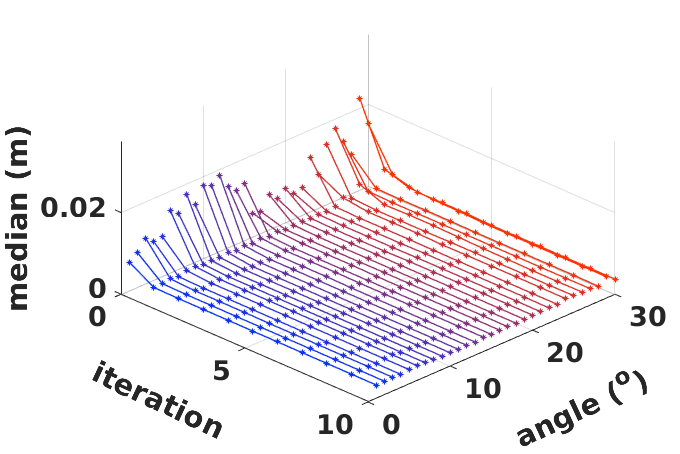}
	
	\caption{Original ICP (\textbf{left}) and our implementation (\textbf{right}) convergence over time (\textbf{top}) and iteration (\textbf{bottom}) for different intra-frame angles.}
	\label{fig:ICP_Angle_Convergences}
	
\end{figure}

In order to show the sensitivity of KinectFusion to angular motion, we considered $20$ consecutive frames from the \textit{Desktop} environment. Between pairs of frames, we rotate the $3$D camera by a specific angle $\theta\in\{5\degree,10\degree,\ldots,60\degree\}$. 
The rationale is to simulates angular shifts in acquisition.
We then feed the data to KinectFusion and inspect the resulting 3D reconstruction for obvious errors such as duplication of scene geometry. 
Figure~\ref{fig:ICP_Angle_Convergences} plots the median of the residual error over iterations, for a predefined set of intra-frame angles.
Clearly, KinectFusion's original ICP cannot cope with orientation changes of more than $20^\circ$ between consecutive frames and shows an increasing degradation. 
In contrast, our IMU-based approach maintains the same convergence quality over angle.

\Table{FailRatioAngles} summarizes the failure rate (averaged across $20$ pairs of frames) for each angular shift during the acquisition. 
Clearly, our IMU-based version of KinectFusion is more robust, since it has a much lower failure rate for most rotations.
Our method degrades more gracefully than classical KinectFusion and total failure only occurs at $\theta=60\degree$,  when the point clouds do not overlap anymore.

\begin{table}[t]
	\centering
	\caption{Failure rate across intra-frame acquisition angles. }		    \resizebox{\columnwidth}{!}{%

	\label{tab:FailRatioAngles}
	\begin{tabular}{r<{\hspace{2pt}}||>{\hspace{-3pt}}c<{\hspace{-3pt}}|>{\hspace{-3pt}}c<{\hspace{-3pt}}|>{\hspace{-3pt}}c<{\hspace{-3pt}}|>{\hspace{-3pt}}c<{\hspace{-3pt}}|>{\hspace{-3pt}}c<{\hspace{-3pt}}|>{\hspace{-3pt}}c<{\hspace{-3pt}}|>{\hspace{-3pt}}c<{\hspace{-3pt}}|>{\hspace{-3pt}}c<{\hspace{-3pt}}|>{\hspace{-3pt}}c<{\hspace{-3pt}}}
		\textbf{\textsl{Desktop}}	& 5\degree & 10\degree & 15\degree & 20\degree & 25\degree & 30\degree & 40\degree & 50\degree & 60\degree \\ \midrule
		\textbf{Original}	& 0\% & 10\% & 25\% & 55\% & 75\% & 90\% & \textbf{100\%} & \textbf{100\%} & \textbf{100\%}   \\
		\textbf{Ours} 		& 0\% & 0\% & 10\% & 0\% & 10\% & 0\% & 5\% & 10\% & \textbf{100\%}  \\ 
	\end{tabular}
	}
\end{table}

\subsection{Improvements in reconstruction}

\begin{figure*}[htb]
	\centering
	\subfloat[][People ($\sim$150 frames)]
	{
		\includegraphics[height=4.1cm]{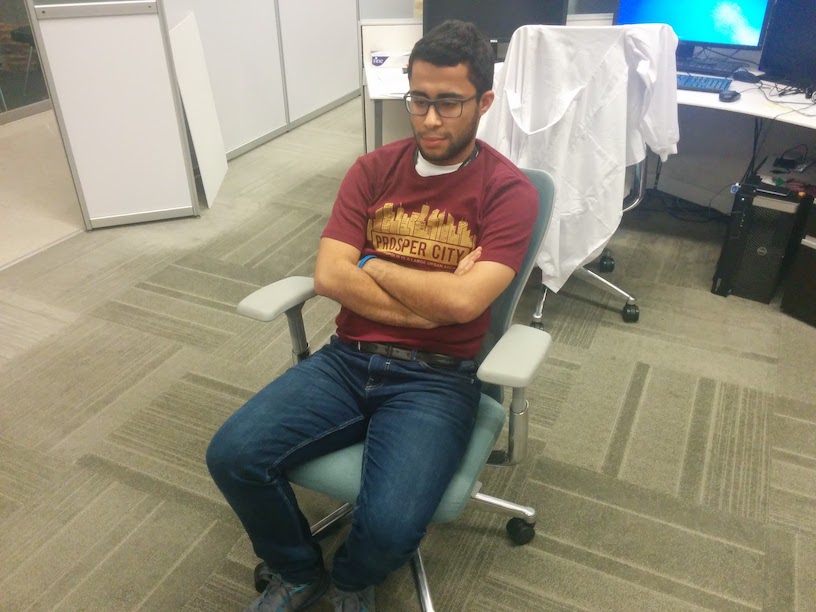}
		\includegraphics[height=4.1cm]{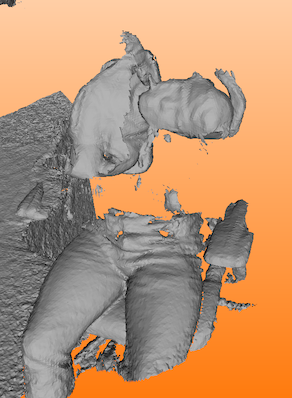}
		\includegraphics[height=4.1cm]{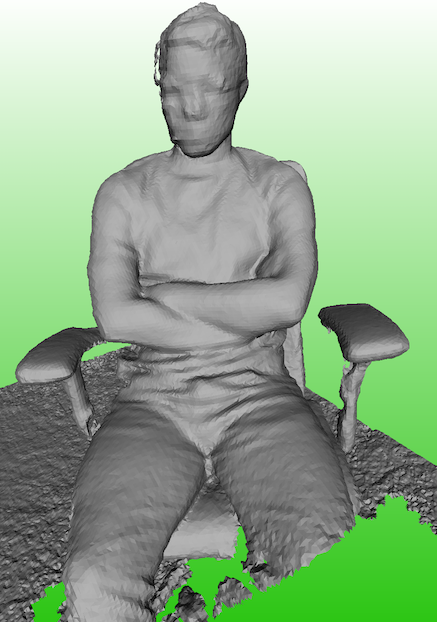}
		\includegraphics[height=4.1cm]{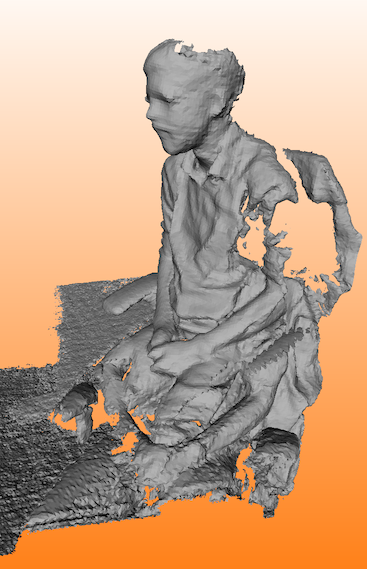}
		\includegraphics[height=4.1cm]{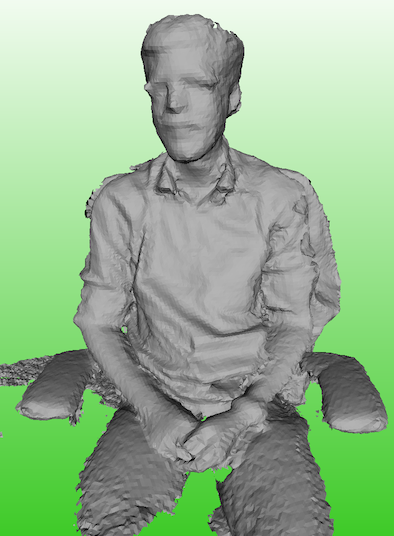}
		\label{fig:ReconstructionPeople}
	}\\
	\subfloat[][{Showcase ($\sim$300 frames)}]	
	{
		\includegraphics[height=3.36cm]{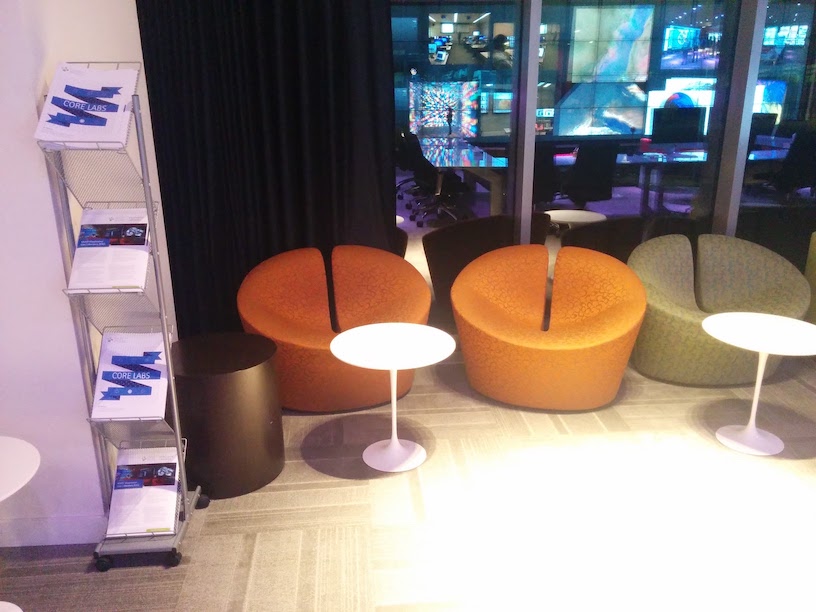}
		\includegraphics[height=3.36cm]{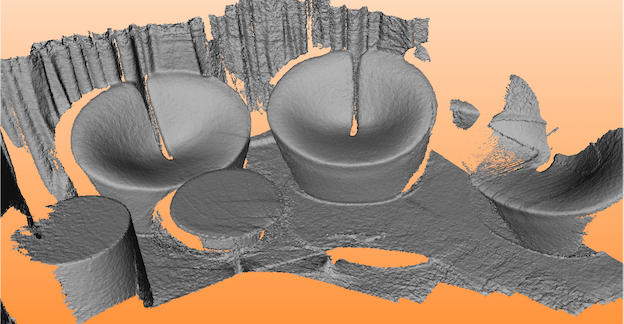}
		\includegraphics[height=3.36cm]{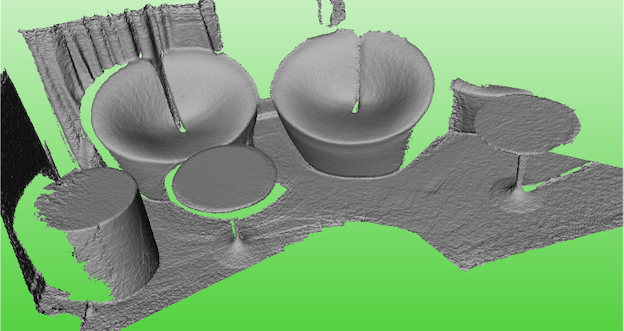}
		\label{fig:ReconstructionShowcase}
	}\\
	\subfloat[][Desktop ($\sim$200 frames)]	
	{
		\includegraphics[height=2.8cm]{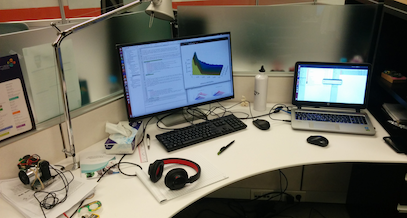}
		\includegraphics[height=2.8cm]{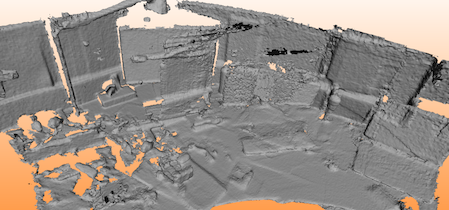}
		\includegraphics[height=2.8cm]{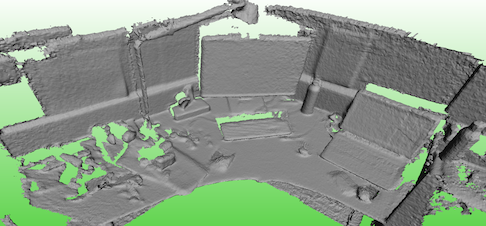}
		\label{fig:ReconstructionDesktop}
	}\\
	\caption{KinectFusion reconstruction of multiple scenes, each with (green background) and without IMU (red background).}
	\label{fig:OtherReconstruction}
\end{figure*}

\begin{figure}[htb]
	\centering
	\subfloat[][ICP convergence for slow motion]
	{\includegraphics[width=0.235\textwidth]{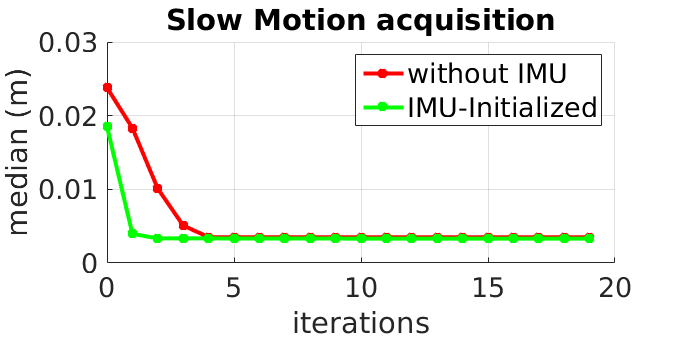}
		\label{subfig:slowmotionregistration}}
	\subfloat[][ICP convergence for fast motion]
	{\includegraphics[width=0.235\textwidth]{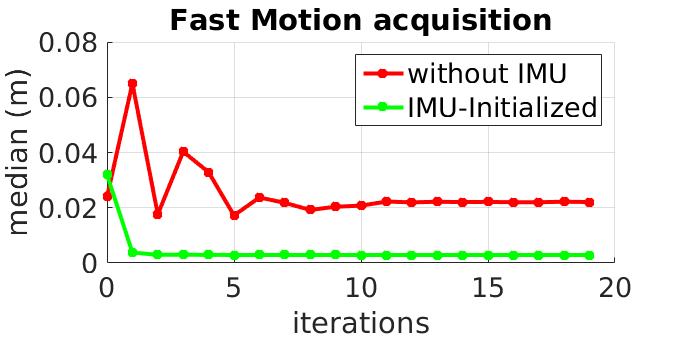}
		\label{subfig:fastmotionregistration}}
	\caption{ICP convergence plot representative for more than $95\%$ of our data, for slow (a) and fast (b) camera motion. Adding an IMU substantially improves both cases.}
    \label{fig:KinFUSpeedConvergence}
\end{figure}

For the same reason we have mentioned above, the reconstruction deteriorates with the scanning pace; a fast scanning at a fixed frame rate creates larger gaps between consecutive frames than a slow scanning.
Figure~\ref{fig:KinFUSpeedConvergence} shows the ICP convergence for a slow and fast pace reconstruction.
Clearly, for slow motion (small angles between consecutive frames), both versions of ICP tend to converge to similar results, with the IMU-based version being faster to converge (refer to Figure~\ref{subfig:slowmotionregistration}). 
However, for fast motion (large angles between consecutive frames), ICP tends not to converge in the 19 iterations at its disposal (refer to Figure~\ref{subfig:fastmotionregistration}).
Because error rampantly propagates, a single instance where acquisition is too fast can lead to a completely erroneous model (refer to Fig.~\ref{fig:Teaser} (red background)). 
In comparison, KinectFusion with IMU-based ICP converges to a much better solution and in a shorter time for the same acquisition (refer to Fig.~\ref{fig:Teaser} (green background)).

In order to assess our hypothesis, we asked different users to perform several reconstructions on multiple scenes with our setup.
To prevent bias, the users were not given intricate instructions about the acquisition strategy they should use.
Note that the users were visualizing the reconstruction as it was being incrementally created. 
As a result, they scanned the scene at a speed that they considered suitable for reconstruction.

Figure~\ref{fig:Teaser} shows some qualitative results for the \textit{Kitchen} dataset, while Figure~\ref{fig:OtherReconstruction} shows results for the \textit{People}, \textit{Showcase} and \textit{Desktop} environments. 
A red background indicates an erroneous reconstruction performed by the original KinectFusion algorithm and a green background shows the reconstruction performed by our implementation.
In these particular examples, we only used the IMU as a seed to the ICP without any regularization, median filtering nor convergence check.
We can already see an improvement in the reconstruction quality.

We notice for the \textit{People} model in Figure~\ref{fig:ReconstructionPeople} that a fast rotation along the camera principal axis (model on the middle) or an horizontal axis (model on the right) creates an inconsistent reconstruction. 
At a certain point, the coarse-to-fine ICP is not able to align the newly acquired point cloud with the model within the 19 iterations at its disposal. 
As a result, it duplicates the scene and overwrites the weight of the TSDF cube that stores the reconstruction.
For the \textit{Showcase} scene in Figure~\ref{fig:ReconstructionShowcase},
a fast motion aligned a frame on the wrong seat, creating a shift in the overall reconstruction, duplicating the objects.
Using the orientation provided by the IMU helped in keeping track of the reconstruction.
For the \textit{Desktop} scene shown in Figure~\ref{fig:ReconstructionDesktop}, a bad alignment has created a duplication of the desktop in two different levels.
The first part of the point cloud streaming reconstructed a model of the desk, but after a bad alignment occurred, a second model was built.

For all the previous examples, our implementation (green background) was able to cope with the difficulties of the different datasets, showing improved robustness in the presence of fast motion.

\subsection{Considerations on convergence timings}

The original KinectFusion method uses a fixed cadence $(4,5,10)$ of coarse-to-fine ICP iterations, since computing the residual error on a GPU is an expensive \textsl{log-reduce} operation.
By feeding the ICP with an initial orientation from the IMU, we can reduce the number of iteration to $(2,2,3)$ without reducing the performance.
Table~\ref{tab:KinFuTiming} shows the improvement in time, which corresponds to 61\%--70\% of the time used for the ICP alone and 35\%--43\% of the overall KinectFusion pipeline.

\begin{table}[htb]
	\centering
	\caption{KinectFusion runtime with custum number of iterations on NVIDIA K6000 and GTX850M GPUs.}		    \resizebox{\columnwidth}{!}{%

	\label{tab:KinFuTiming}
	\begin{tabular}{l||c|c||c|c}
		& \multicolumn{2}{c||}{\textbf{K6000}}							& \multicolumn{2}{c}{\textbf{GTX850M}}						\\
		& \textbf{ICP }	&\textbf{KinFu } & \textbf{ICP }	& \textbf{KinFu }  \\ \midrule
		\textbf{Original (4,5,10)}	& $10.58$~ms 	& $15.74$~ms 	& $27.17$~ms 	& $43.41$~ms \\ 
		\textbf{Ours (2,2,3)} 	& $~4.07$~ms 	& $10.22$~ms 	& $~8.22$~ms 	& $24.72$~ms  \\ \midrule
		\textbf{\textit{Improvement}}	& \textit{$-61.53~\%$} 	& \textit{$-35.07~\%$} 	& \textit{$-69.75~\%$} & \textit{$-43.05~\%$} 	\\ 
	\end{tabular}
	}
\end{table}

\begin{figure}[htb]
	\centering
	\includegraphics[width=0.5\textwidth]
	{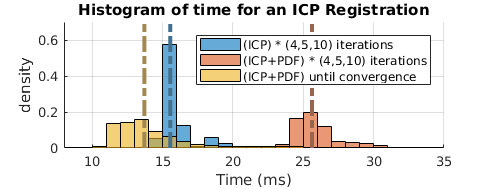}
	\caption{Computational time for the original KinectFusion (blue), KinectFusion with PDF estimation for distant point removal (red) and KinectFusion with PDF estimation and median value convergence check (yellow).}
	\label{fig:KinFu_Time_Evalutation}
\end{figure}

Since less ICP iterations result in improved speed, we can use this saved time  to estimate the PDF of the closest-point distances. Such an operation can be time consuming if performed on the CPU, so we take advantage of the multiple GPU cores available to build the distance PDF on the GPU as well.
Figure~\ref{fig:KinFu_Time_Evalutation} shows the timing for different versions of  KinectFusion, while reconstructing the Freiburg scenes (a total of 15,793 alignments). In blue, we show the distribution of the time required by the original KinectFusion, with an average of $15.55$ms. Adding the estimation of the PDF at each ICP iteration (red distribution), the average time increases to $25.62$ms. However, once we use the PDF to estimate its median value and stop the alignment after a convergence is detected, our method requires only $11.54$ iterations on average to converge, whereas the original KinectFusion requires $19$ iterations. Since our method converges in less iterations, it only requires an overall average of $13.71$ms for alignment, which corresponds to an improvement of $11.8$\% on the original time. Furthermore, Table~\ref{tab:KinFu_Time_Evalutation} shows the timing details for different datasets.

\begin{table}[htb]
	\centering
	\caption{KinectFusion runtime for the \textit{original} KinectFusion, our version using a \textit{fixed} number of iteration (4,5,10) and our version with a control of the \textit{convergence}.}
		    \resizebox{\columnwidth}{!}{%

	\label{tab:KinFu_Time_Evalutation}
	\begin{filecontents*}{img/timing/kinfuconvergencetiming.csv}
Dataset,NbPC,MeanOriginal,MeanFixed,MeanConvergence,MeanImprovement,MedianOriginal,MedianFixed,MedianConvergence,MedianImprovement
fr1/desk,571,15.968476357268,26.584938704028,12.8336252189142,-19.631498135556,15,26,13,-13.3333333333333
fr1/desk2,609,16.1271676300578,25.8095238095238,14.0985221674877,-12.5790560940728,15,25,13,-13.3333333333333
fr1/plant,1117,15.3647798742138,25.390243902439,14.8845360824742,-3.12561452667204,15,25,14,-6.66666666666667
fr1/room,1351,15.7780252412769,25.6922357343311,13.8341968911917,-12.3198456103363,15,25,13,-13.3333333333333
fr1/rpy,689,15.4078374455733,25.8272859216255,12.8984034833091,-16.2867370007536,15,25,12,-20
fr1/teddy,1398,15.3767441860465,25.1314168377823,13.0594059405941,-15.070409037274,15,24,13,-13.3333333333333
fr1/xyz,789,15.7566539923954,26.3536121673004,13.084917617237,-16.956241956242,15,26,13,-13.3333333333333
fr2/desk,2173,15.6566958122411,25.7898916967509,14.416014726185,-7.92428428663806,15,25,13,-13.3333333333333
fr2/dishes,2940,15.2579755809374,25.0825531914894,14.2198815009875,-6.80361607896942,15,24,13,-13.3333333333333
fr2/ms2,1834,15.5648604269294,25.5104166666667,13.0986547085202,-15.8447017882816,15,25,12,-20
fr3/teddy,2322,15.5526167698368,25.2669245647969,12.7461368653422,-18.0450656376849,15,25,12,-20
AVG,15793,15.5538563340931,25.6197530864198,13.7119872701556,-11.8418803952833,15,25,13,-13.3333333333333
\end{filecontents*}
\csvreader[tabular=l|r|r|r|r|r, 
	table head=  &	 &  \bfseries \textit{Original} &  \bfseries \textit{Fixed} & \bfseries \textit{Until} & \bfseries \textit{Impro-}\\
	\bfseries Dataset &	\bfseries nPC  &  \bfseries \textit{KinFu} &  \bfseries \textit{iteration} & \bfseries \textit{Converg.} & \bfseries \textit{vements}\\\midrule,
	late after line=\ifthenelse{\equal{\Dataset}{AVG}}{\\\midrule}{\\}]
	{img/timing/kinfuconvergencetiming.csv}%
	{Dataset=\Dataset,NbPC=\NbPC,MeanOriginal=\MeanOriginal,MeanFixed=\MeanFixed,MeanConvergence=\MeanConvergence,MeanImprovement=\MeanImprovement}%
	{\textbf{\Dataset} & \NbPC & 
		\num[round-mode=places,round-precision=1]{\MeanOriginal}~ms & 
		\num[round-mode=places,round-precision=1]{\MeanFixed}~ms & 
		\num[round-mode=places,round-precision=1]{\MeanConvergence}~ms & 
		\num[round-mode=places,round-precision=1]{\MeanImprovement}~\%
		}
		}
\end{table}

\section{Performance on the Freiburg Benchmark} \label{sec:Dataset}

We evaluate our implementation on the Freiburg dataset~\cite{sturm2012benchmark}, which is commonly used to benchmark  large-scale SLAM techniques. Since this dataset does not provide any measured orientation, we first characterize the uncertainty model of our IMU in order to apply such a model to the ground truth orientation.

\subsection{Metrological Characterization of the IMU Orientation}


To model the IMU noise and include it in the Freiburg dataset, we characterize the uncertainty of the IMU under static conditions according the Guide to the Expression of Uncertainty in Measurement (GUM)~\cite{bipm2008evaluation}.
We used a $6$-DoF anthropomorphic robot ABB IRB $1200$ to stress the IMU in rotation along the three main axes, within a $\pm180\degree$ range and a $5\degree$ angular step. 
The reference orientation has a $0.01\degree$ resolution on each axis, which is several orders of magnitude better in resolution than the expected IMU uncertainty.

For the static characterization, the IMU was positioned in space with a controlled orientation with respect to its reference system (Earth). The \Z-axis vertical was aligned to \textsl{top}, and the \X- and \Y- axes orthogonal and horizontal. In particular, the \X- axis was aligned with the \textsl{north} direction.
After a stabilization time of 5 seconds, $1000$ orientation measurements are performed at the IMU frequence of $100$Hz. Uncertainty was measured as a combination of a \emph{random error} represented by the standard deviation of the $1000$ measurements and a \emph{systematic error}  measured by comparison to the ground truth orientation provided by the robotic arm.

\begin{figure}[htb]
	\centering
	
	    \includegraphics[width=0.235\textwidth]
	    {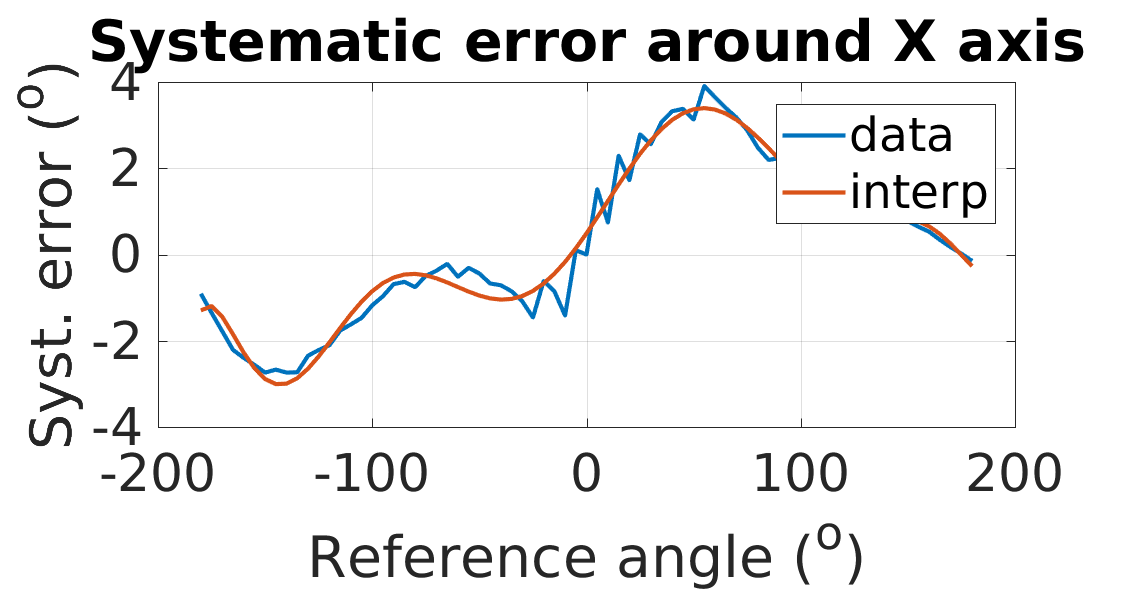}
	    \includegraphics[width=0.235\textwidth]
	    {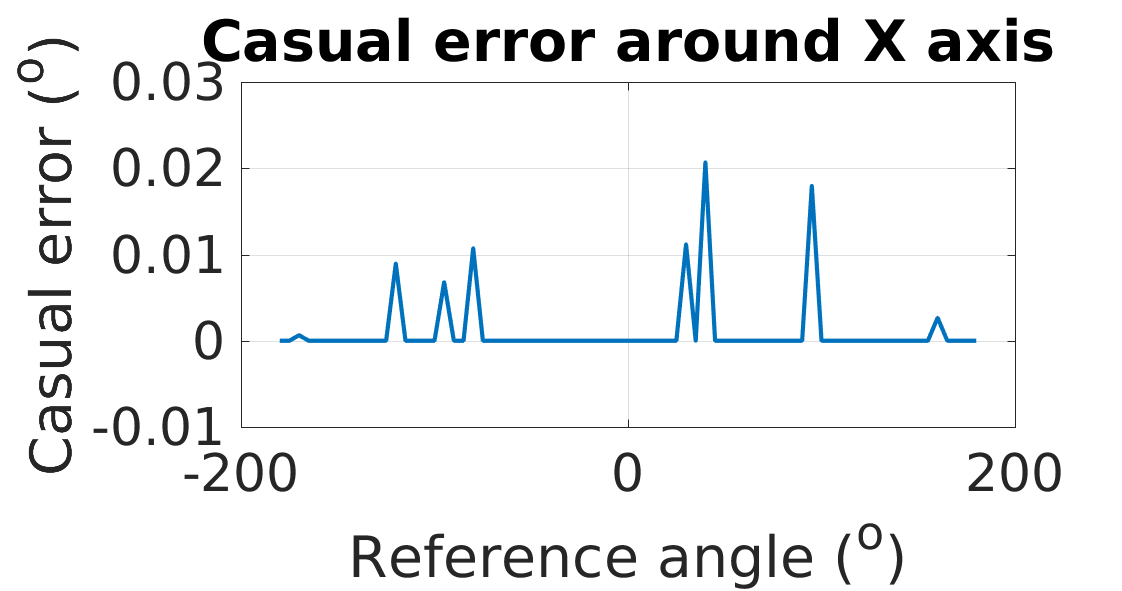}\\
	    \includegraphics[width=0.235\textwidth]
	    {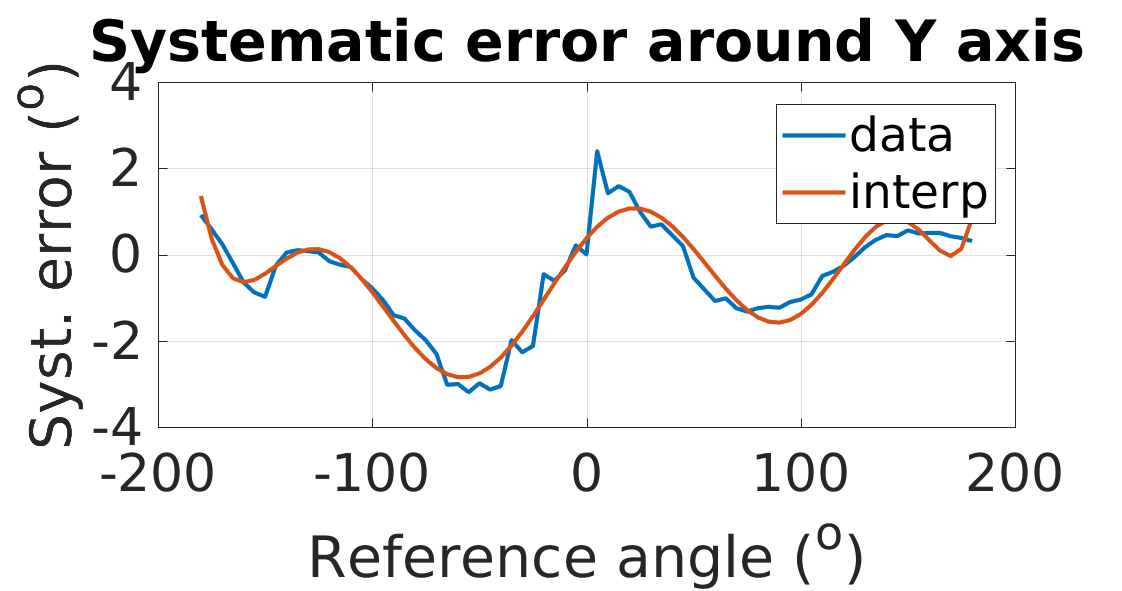}
	    \includegraphics[width=0.235\textwidth]
	    {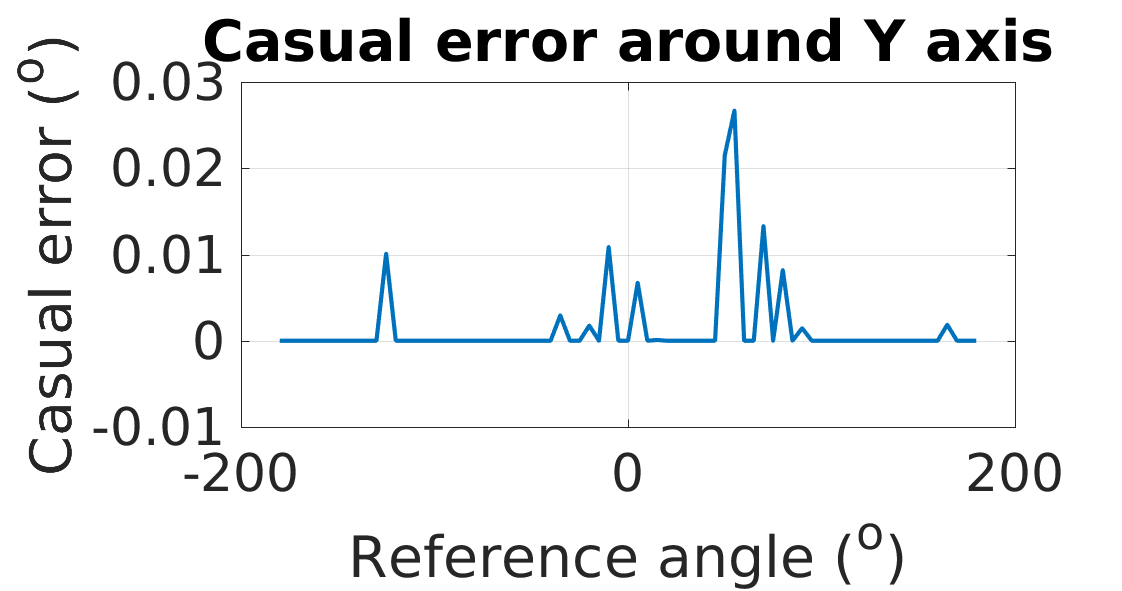}\\
	    \includegraphics[width=0.235\textwidth]
	    {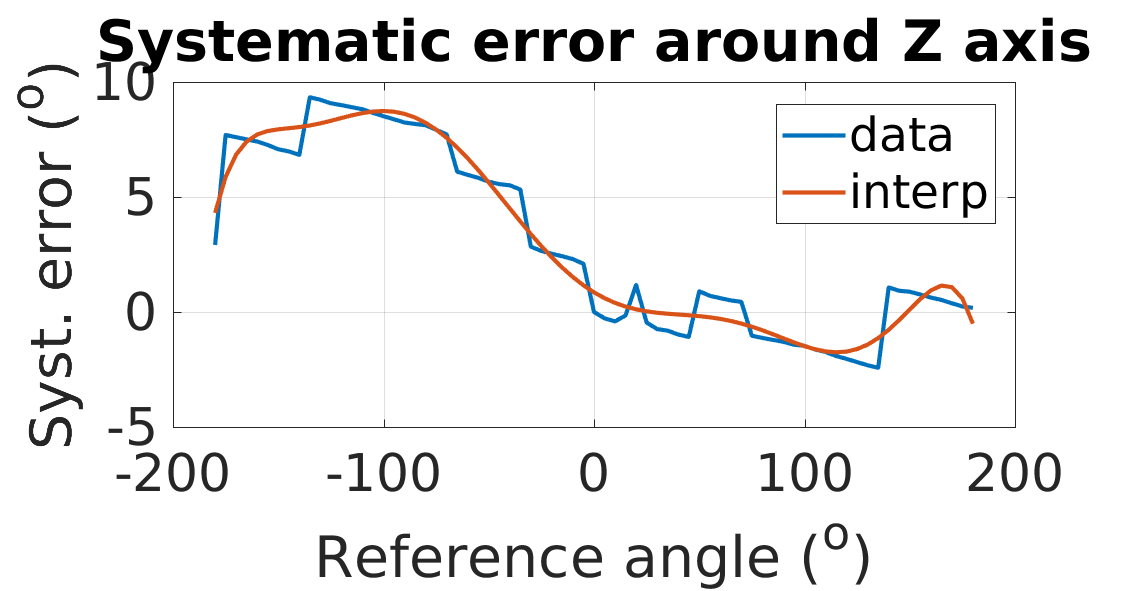} 
	    \includegraphics[width=0.235\textwidth]
	    {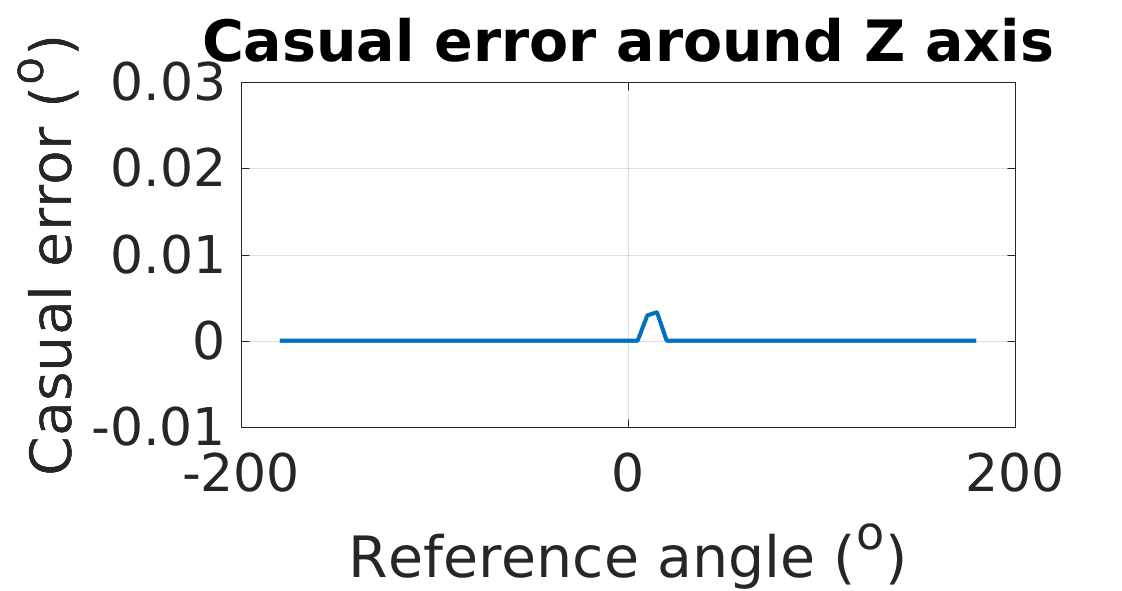}\\
	
	\caption{Uncertainty characterization of the BNO055 IMU along \X-, \Y- and \Z- axes for the systematic (top) and random (bottom) errors. Note the different scales.}
	\label{fig:UncertaintyCharacterization}
\end{figure}

Results presented in Figure~\ref{fig:UncertaintyCharacterization} show that the uncertainty has an important systematic error, but no random error.
The absence of a random component in the uncertainty is due to a strong low pass filter occurring on the IMU for the orientation estimation. 
The systematic error is mainly due to the internal calibration between the sensors.
For reference, the IMU orientation around the horizontal \X- and \Y- axes are extracted using the combination of a magnetometer, an accelerometer, and a gyroscope, which provide accuracy around $\pm3\degree$. Orientation around the vertical \Z-axis is extracted using the magnetometer only, which provides lower quality measurement with uncertainty measured to be about $\pm10\degree$. The test was  repeated several times with different initial orientations for the IMU leading to similar trends.

\subsection{Methods and Results}

We used multiple datasets from the Freiburg benchmark~\cite{sturm2012benchmark}, focusing on 3D object reconstructions rather than large environments. The list of datasets we used is shown in Table~\ref{tab:ATERPE}.
We adapted the parameters of the KinectFusion such that the reconstructions can fit inside a $512\times512\times512$ TSDF cube, with a maximum side length of $6$m.
We chose identical parameters to run instances of the original KinectFusion and our pipeline using the ground truth orientation, with and without the systematic error that we characterized in the uncertainty model.

We tested our model over different values of $\lambda$ to figure out the best regularization. 
\Table{Regularization} shows the average improvement for Absolute Trajectory Error (ATE) metrics (in \%) with respect to the original KinectFusion algorithm, over all the datasets used.
Our results clearly indicate that larger regularization results in better accuracy.

\begin{table}[htb]
	\centering
	\caption{Comparison of the average RMSE Absolute Trajectory Error (ATE) of our method respect to the original KinectFusion one using different regularizations parameter $\lambda$.
	Values are given in percentage of improvement over all the datasets.}
	\resizebox{\columnwidth}{!}{%

	\label{tab:Regularization}
	\begin{filecontents*}{img/freiburg/ate_regularization.csv}
pippo,Cone,Ctwo,Cthree,Cfour,Cfive,AVERAGE
$c$,-29.91568789,4.661220689,-28.93852639,-33.31487678,-43.68989455,-26.23955298
$c / \sqrt{n}$,-26.2291526,-31.29484938,-32.84374016,-46.45088418,-45.51304094,-36.46633345
$c/n$,-6.340281355,-10.59902937,-51.00013569,-40.78152209,-28.89552197,-27.5232981
$c/n^2$,20.30534096,-2.287778634,-44.63362218,-33.89520116,-40.56914402,-20.216081
-$c~\text{ln}(n)$,72.06227272,-32.10671921,-16.14736686,-33.67480673,-41.12075834,-10.19747569
AVG,5.976498366,-14.32543118,-34.71267826,-37.62345819,-39.95767196,-24.12854824
\end{filecontents*}
\csvreader[tabular = r||r|r|r|r|r||r, 
	table head=  \textbf{c~=}	& \textbf{0.05}	& \textbf{0.20}	& \textbf{1.00}	& \textbf{2.00}	& \textbf{5.00} & \textbf{AVG}\\\midrule,
	late after line=\ifthenelse{\equal{\pippo}{AVG}}{\\\midrule}{\\}]
	{img/freiburg/ate_regularization.csv}%
	{pippo=\pippo,
		Cone=\Cone,
		Ctwo=\Ctwo,
		Cthree=\Cthree,
		Cfour=\Cfour,
		Cfive=\Cfive,
		AVERAGE=\AVERAGE}%
	{\pippo & 
		\num[round-mode=places,round-precision=0]{\Cone}\% &
		\num[round-mode=places,round-precision=0]{\Ctwo}\% &
		\num[round-mode=places,round-precision=0]{\Cthree}\% &
		\num[round-mode=places,round-precision=0]{\Cfour}\% &
		\num[round-mode=places,round-precision=0]{\Cfive}\% &
		\num[round-mode=places,round-precision=0]{\AVERAGE}\%}
		}
\end{table}

\begin{figure}[!htb]
	\centering	
	
	\includegraphics[width=0.113\textwidth]	{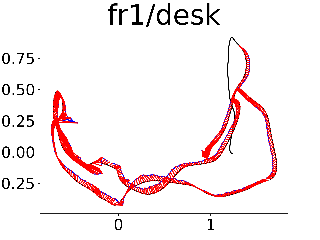}
	\includegraphics[width=0.113\textwidth]	{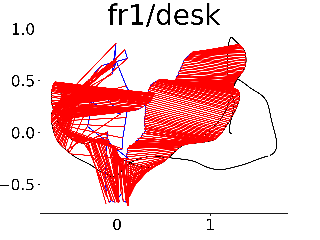}
	\includegraphics[width=0.113\textwidth]	{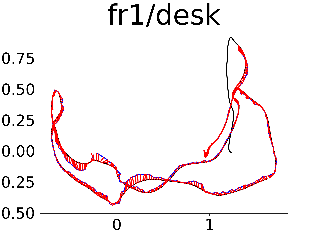}
	\includegraphics[width=0.113\textwidth]	{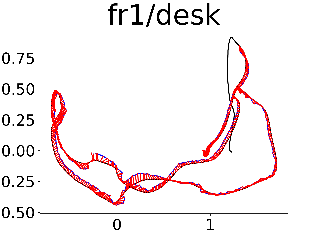}

	\includegraphics[width=0.113\textwidth]	{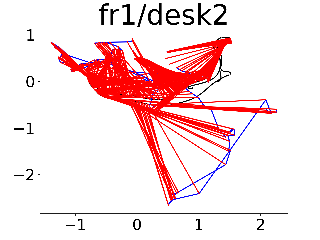}
	\includegraphics[width=0.113\textwidth]	{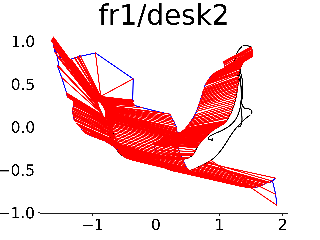}
	\includegraphics[width=0.113\textwidth]	{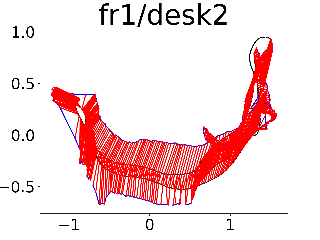}	
	\includegraphics[width=0.113\textwidth]	{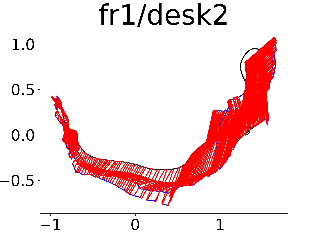}

	\includegraphics[width=0.113\textwidth]	{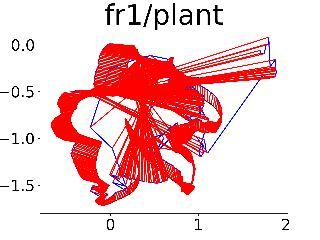}
	\includegraphics[width=0.113\textwidth]	{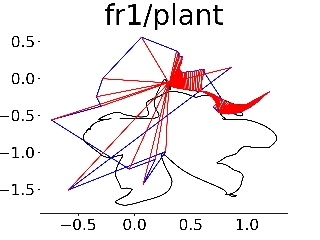}
	\includegraphics[width=0.113\textwidth]	{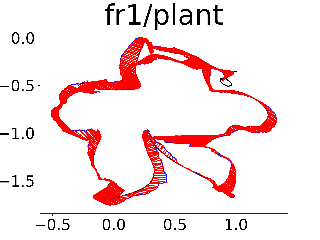}
	\includegraphics[width=0.113\textwidth]	{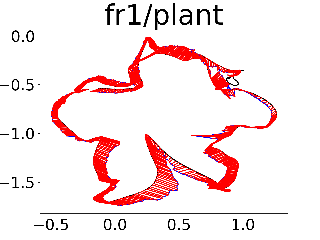}

	\includegraphics[width=0.113\textwidth]	{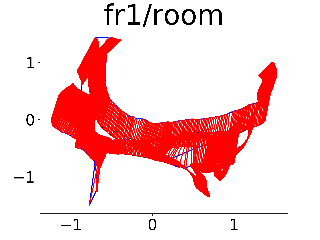}
	\includegraphics[width=0.113\textwidth]	{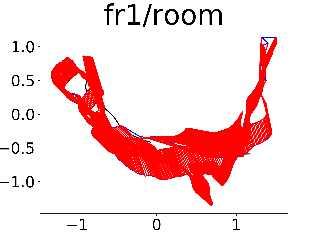}
	\includegraphics[width=0.113\textwidth]	{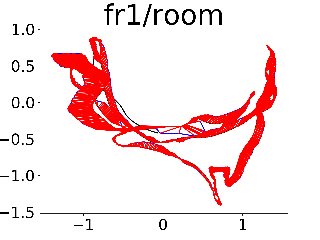}	
	\includegraphics[width=0.113\textwidth]	{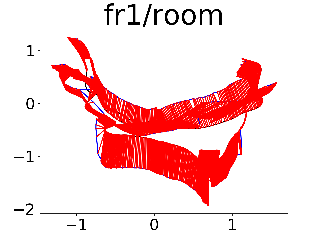}
	
	\includegraphics[width=0.113\textwidth]	{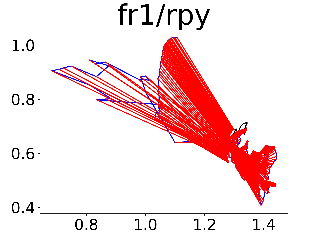}
	\includegraphics[width=0.113\textwidth]	{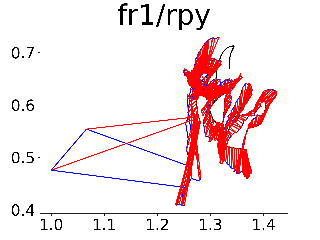}
	\includegraphics[width=0.113\textwidth]	{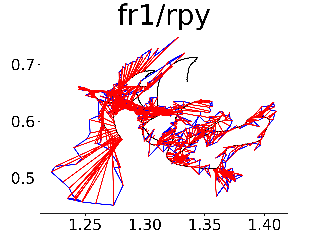}	
	\includegraphics[width=0.113\textwidth]	{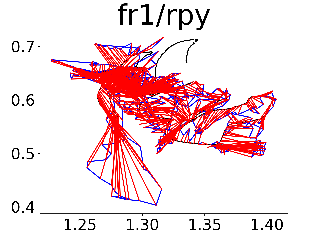}
	
	\includegraphics[width=0.113\textwidth]	{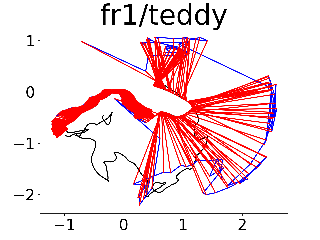}
	\includegraphics[width=0.113\textwidth]	{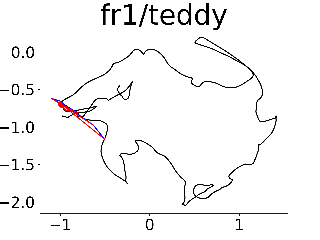}
	\includegraphics[width=0.113\textwidth]	{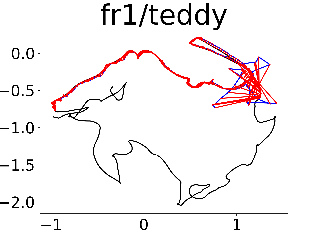}	
	\includegraphics[width=0.113\textwidth]	{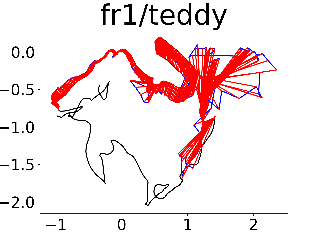}
	
	\includegraphics[width=0.113\textwidth]	{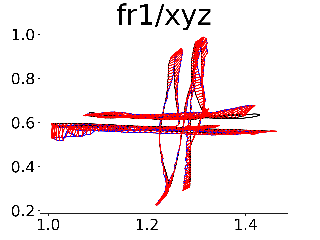}
	\includegraphics[width=0.113\textwidth]	{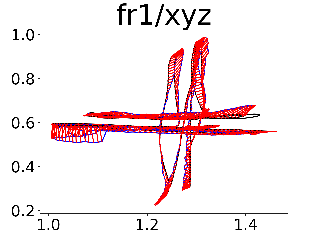}
	\includegraphics[width=0.113\textwidth]	{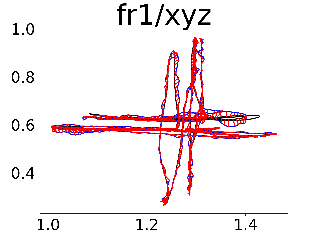}	
	\includegraphics[width=0.113\textwidth]	{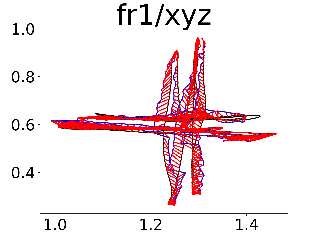}
	
	\includegraphics[width=0.113\textwidth]	{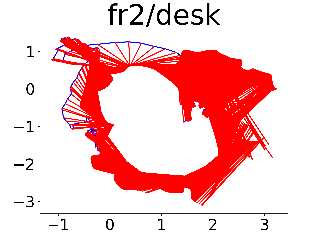}
	\includegraphics[width=0.113\textwidth]	{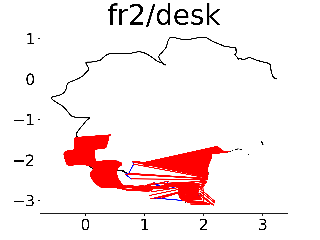}
	\includegraphics[width=0.113\textwidth]	{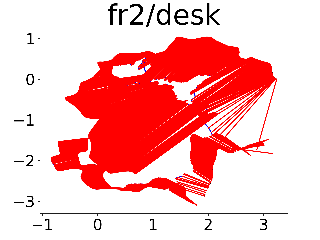}	
	\includegraphics[width=0.113\textwidth]	{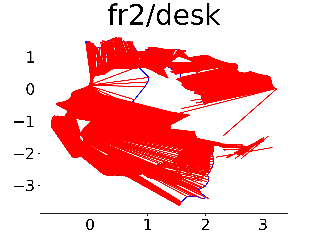}

	\includegraphics[width=0.113\textwidth]	{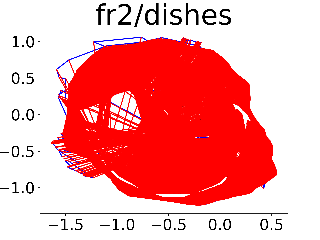}
	\includegraphics[width=0.113\textwidth]	{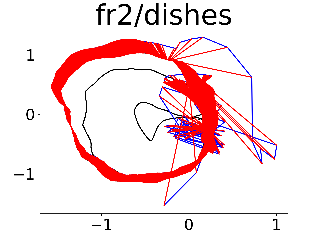}
	\includegraphics[width=0.113\textwidth]	{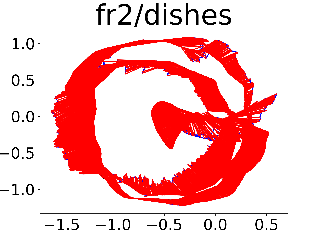}	
	\includegraphics[width=0.113\textwidth]	{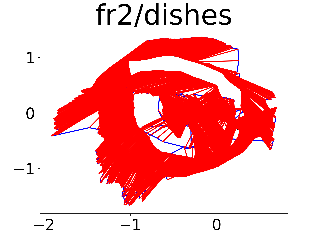}

	\includegraphics[width=0.113\textwidth]	{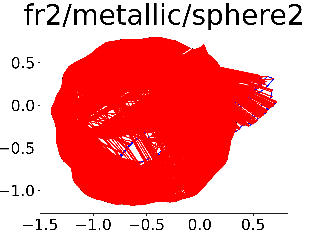}
	\includegraphics[width=0.113\textwidth]	{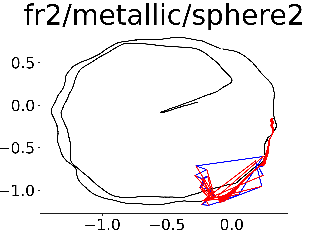}
	\includegraphics[width=0.113\textwidth]	{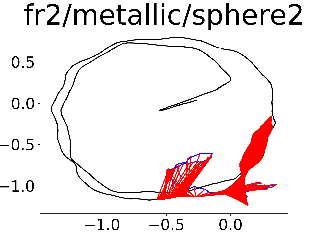}	
	\includegraphics[width=0.113\textwidth]	{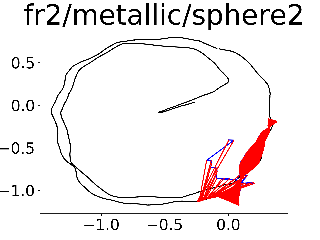}

	\includegraphics[width=0.113\textwidth]	{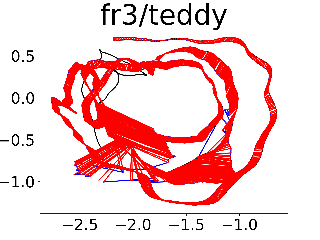}
	\includegraphics[width=0.113\textwidth]	{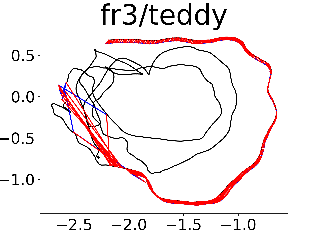}
	\includegraphics[width=0.113\textwidth]	{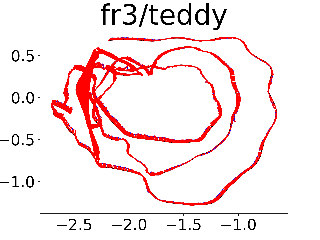}	
	\includegraphics[width=0.113\textwidth]	{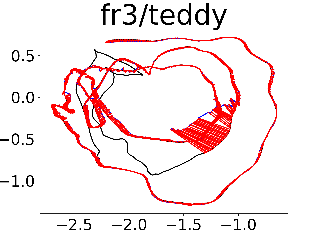}

	\caption{
		\textbf{From top to bottom:} Trajectories for some Freiburg datasets. 
		(\textbf{black:} ground truth, \textbf{blue:} estimated, \textbf{red:} difference)
		\textbf{From left to right:} Original KinectFusion / 
		our method without regularization ($\lambda=0$) / 
		adding regularization ($\lambda=5$) and convergence check / 
		using noisy orientation.  }
	\vspace{-2mm}

	\label{fig:OURSprevious}
\end{figure}

Table~\ref{tab:ATERPE} details the ATE and the Relative Pose Estimation (RPE) metrics for the different datasets.
We only show the improvement on the translational part, since the rotational part is directly measured.
We can see that the average ATE error is reduced by $53\%$ on the selected dataset and the RPE is reduced by $21\%$. Figure~\ref{fig:OURSprevious} plots the trajectories resulting from our method. They are consistent with the ground truth. When using the original KinectFusion method, part of the trajectory is not reconstructed due to tracking loss.

\begin{table}[htb]
	\centering
	\caption{ATE and RPE metrics for our method compared to the original KinectFusion. Here we used regularization ($\lambda=5$), median filtering and convergence control }
	    \resizebox{\columnwidth}{!}{%

	\label{tab:ATERPE}
	\begin{filecontents*}{img/freiburg/ate_rpe_noise.csv}
Dataset,ATEorig,ATEours,ATEoursnoise,RPEorig,RPEours,RPEoursnoise
fr1/desk,0.072962,0.030087,0.044159,0.738048,0.74483,0.739424
fr1/desk2,1.161819,0.292637,0.444369,1.287929,0.838171,0.869576
fr1/plant,0.569372,0.134107,0.176352,0.706834,0.654759,0.632407
fr1/room,0.533064,0.29157,0.369766,0.572936,0.568497,0.57648
fr1/rpy,0.143752,0.02627,0.039267,0.189792,0.147253,0.155186
fr1/teddy,0.771908,0.032981,0.136453,1.107038,0.562483,0.551832
fr1/xyz,0.02201,0.017639,0.031814,0.471017,0.455407,0.44321
fr2/desk,1.398591,1.086605,1.106806,0.513134,0.325256,0.411005
fr2/dishes,1.117713,0.26825,0.619345,0.39359,0.239472,0.323013
fr2/ms2,1.126076,0.140761,0.142085,0.426254,0.321132,0.329455
fr3/teddy,0.396014,0.097218,0.292001,0.430286,0.392398,0.397371
AVG,0.664843727272727,0.219829545454545,0.309310636363636,0.621532545454545,0.477241636363636,0.493541727272727
Improv.,100,-66.9351553700726,-53.4761894148468,100,-23.2153424862707,-20.5927781445804
\end{filecontents*}
\csvreader[tabular=l||r|r|r||r|r|r, 
	table head=      & \multicolumn{3}{c||}{\textbf{ATE (RMSE) (m)}} & \multicolumn{3}{c}{\textbf{RPE (RMSE) (m)}}\\
	\textbf{dataset} &    & \multicolumn{2}{c||}{Ours} &       & \multicolumn{2}{c}{Ours}  \\
	            	& Orig. & IMU &  +noise & Orig. & IMU &  +noise \\\midrule,
	late after line=\ifthenelse{\equal{\Dataset}{AVG}}{\\\midrule}{\\}]
	{img/freiburg/ate_rpe_noise.csv}%
	{Dataset=\Dataset,
		ATEorig=\ATEorig,
		ATEours=\ATEours,
		ATEoursnoise=\ATEoursnoise,
		RPEorig=\RPEorig,
		RPEours=\RPEours,
		RPEoursnoise=\RPEoursnoise}%
	{\ifthenelse{\equal{\Dataset}{Improv.}}
		{\textbf{\Dataset}}
		{\textbf{\Dataset}} & 
		\ifthenelse{\equal{\Dataset}{Improv.}}
		{-}
		{\num[round-mode=places,round-precision=3]{\ATEorig}} & 
		\ifthenelse{\equal{\Dataset}{Improv.}}
		{\num[round-mode=places,round-precision=0]{\ATEours}\%}
		{\num[round-mode=places,round-precision=3]{\ATEours}} & 
		\ifthenelse{\equal{\Dataset}{Improv.}}
		{\num[round-mode=places,round-precision=0]{\ATEoursnoise}\%}
		{\num[round-mode=places,round-precision=3]{\ATEoursnoise}} & 
		\ifthenelse{\equal{\Dataset}{Improv.}}
		{-}
		{\num[round-mode=places,round-precision=3]{\RPEorig}} & 
		\ifthenelse{\equal{\Dataset}{Improv.}}
		{\num[round-mode=places,round-precision=0]{\RPEours}\%}
		{\num[round-mode=places,round-precision=3]{\RPEours}} & 
		\ifthenelse{\equal{\Dataset}{Improv.}}
		{\num[round-mode=places,round-precision=0]{\RPEoursnoise}\%}
		{\num[round-mode=places,round-precision=3]{\RPEoursnoise}}
	}
	}
\end{table}

\section{Conclusion} \label{sec:Conclusion}

In this work, we have demonstrated the benefits of integrating a cheap and modular IMU into the popular KinectFusion reconstruction pipeline. 
We used the IMU orientation to seed the ICP algorithm, which allows us to linearize/convexify the original problem more faithfully. 
We used a regularized point-to-plane metric that constrains the orientation within boundaries. 
We made sure the chosen correspondences are consistent and free of outliers by exploiting their median distance as a basis for outlier removal. 
We showed qualitative and quantitative improvements in the robustness of our modified KinectFusion pipeline over the original KinectFusion.
In addition to improved reconstruction quality, speed is also improved by almost 12\%. This is due to a significant reduction in the number of ICP iterations needed for convergence.
Finally, results on the Freiburg benchmark show that the overall quality of the tracking/reconstruction is improved by a factor of $2$.

\textbf{Acknowledgments.} This work was supported by the King Abdullah University of Science and Technology (KAUST) Office of Sponsored Research and the Visual Computing Center (VCC).


{\footnotesize
\bibliographystyle{ieee}
\bibliography{biblio}
}

\end{document}